\documentclass[AMA,LATO1COL]{WileyNJD-v2}
\usepackage{subcaption}
\usepackage{float}

\articletype{Article Type}%

\received{26 April 2016}
\revised{6 June 2016}
\accepted{6 June 2016}

\begin{document}

\title{Fast PDN Impedance Prediction Using Deep Learning}

\author[1]{Ling Zhang}

\author[1]{Jack Juang}

\author[1]{Zurab Kiguradze}

\author[1]{Bo Pu}

\author[2]{Shuai Jin}

\author[2]{Songping Wu}

\author[2]{Zhiping Yang}

\author[1]{Chulsoon Hwang*}

\authormark{Ling Zhang \textsc{et al}}

\address[1]{\orgdiv{EMC Laboratory}, \orgname{Missouri University of Science and Technology}, \orgaddress{\state{Rolla, MO}, \country{USA}}}

\address[2]{\orgdiv{Google Inc.}, \orgaddress{\state{Mountain View, CA}, \country{USA}}}

\corres{*Chulsoon Hwang, EMC LAB, Missouri Univeristy of Science and Technology. \email{hwangc@mst.edu}}

\presentaddress{4000 Enterprise Dr., Rolla, MO 65409}

\abstract[Abstract]{Modeling and simulating a power distribution network (PDN) for printed circuit boards (PCBs) with irregular board shapes and multi-layer stackup is computationally inefficient using full-wave simulations. This paper presents a new concept of using deep learning for PDN impedance prediction. A boundary element method (BEM) is applied to efficiently calculate the impedance for arbitrary board shape and stackup. Then over one million boards with different shapes, stackup, IC location, and decap placement are randomly generated to train a deep neural network (DNN). The trained DNN can predict the impedance accurately for new board configurations that have not been used for training. The consumed time using the trained DNN is only 0.1 seconds, which is over 100 times faster than the BEM method and 5000 times faster than full-wave simulations.}


\maketitle

\section{Introduction}\label{sec1}

Accurate and fast modeling for multi-layer printed circuit boards (PCBs) is of critical importance to the design and performance evaluation of the power distribution network (PDN). Different methodologies have been proposed to model PDN structure and compute impedance \cite{[1]} \cite{[2]} \cite{[3]} \cite{[4]} \cite{[5]}. The cavity model method \cite{[1]} \cite{[2]} is an efficient approach to calculate PDN impedance, but it is limited to rectangular board shapes. The plane-pair PEEC (PPP) method \cite{[3]} can address irregular board shapes but requires solving a 2D mesh circuit and is therefore computationally intensive. There are also some boundary integral methods \cite{[4]} \cite{[5]} that only require 1D integration but are still not efficient enough in some applications. For example, in the pre-layout stage, a substantial amount of computations are needed to optimize design parameters.

In recent years, the success of deep learning for complex and non-linear problems like computer vision \cite{[6]}, natural language processing \cite{[7]}, and strategy games \cite{[8]} have also impacted many other fields. There has been some research \cite{[9]} \cite{[10]} \cite{[11]} \cite{[12]} in applying machine learning in PDN modeling and optimization. However, most of these works do not have a well trained and generalized machine learning model for PDN impedance prediction at the PCB level. In the work of \cite{[9]}, an artificial neural network (ANN) has been adopted to predict target impedance violations for PDN by considering the variations of IC location, decap placement, and target impedance. However, their task is just a simple classification problem to judge if the target impedance will be violated or not. It cannot provide quantitative and insightful details about the actual impedance curve. Moreover, the variation of board shape and stackup is not considered, which makes the trained deep neural network (DNN) hard to generalize.

In this paper, deep learning will be utilized to predict the impedance curve for any board shape, stackup, IC location, and decap placement. A DNN can be trained by using a considerable number of boards with different configuration parameters. Compared to the traditional ways of calculating PDN impedance \cite{[1]} \cite{[2]} \cite{[3]} \cite{[4]} \cite{[5]}, the trained DNN can be much faster while retaining a tolerable accuracy. Therefore, it can be a particularly powerful and efficient tool for PDN impedance evaluation at the design stage.

\begin{figure}[htb]
\centerline{\includegraphics[width=500pt,height=103pt]{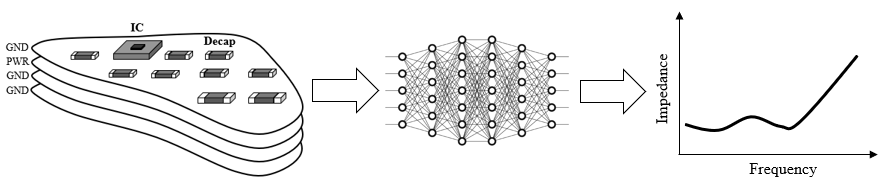}}
\caption{A deep neural network (DNN) can be trained to predict the PDN impedance for different design parameters including board shape, stackup, IC location, and decap placement. \label{fig1}}
\end{figure}

The remaining sections are organized as follows. In section~\ref{sec2}, the impedance calculation method for arbitrary board shape and stackup is briefly introduced, and the data generation process is elaborated. Section~\ref{sec3} shows the detailed DNN structure and the training process. Section~\ref{sec4} demonstrates the testing result for the trained DNN. Finally, conclusions are drawn in section~\ref{sec5}.

\section{Training Data Generation}\label{sec2}

\subsection{Impedance Calculation}

To ensure the good performance of a DNN, abundant board data with different configurations need to be generated for training. Hence, developing an efficient method to calculate the impedance for arbitrary board shape and stackup is crucial to the feasibility of the deep learning algorithm. Consequently, a boundary element method (BEM)  \cite{[5]}  \cite{[13]} that can handle arbitrarily-shaped parallel planes is applied to calculate the quasi-static inductances between vertical vias. In this BEM method, only the boundary needs to be discretized into a proper number of segments for 1D integration. Afterward, an equivalent circuit can be formed by the inductances and parallel-plate capacitances for multi-layer PDN structures. Instead of using commercial tools, the well-known node voltage method is applied to obtain the Z-parameters of the network looking into the IC and decap ports \cite{[13]}. The Z-parameters of decaps can be further connected to the decap ports to obtain the total impedance looking into the IC.

\begin{figure}[htb]
\begin{subfigure}{.33\textwidth}
  \centerline{\includegraphics[width=150pt,height=119pt]{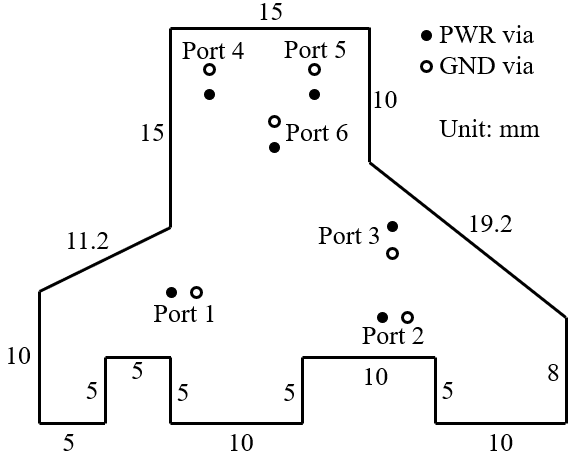}}
  \caption{}
  \label{fig2a}
\end{subfigure}
\begin{subfigure}{.33\textwidth}
  \centerline{\includegraphics[width=139pt,height=60pt]{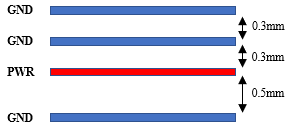}}
  \caption{}
  \label{fig2b}
\end{subfigure}
\begin{subfigure}{.33\textwidth}
  \centerline{\includegraphics[width=159pt,height=119pt]{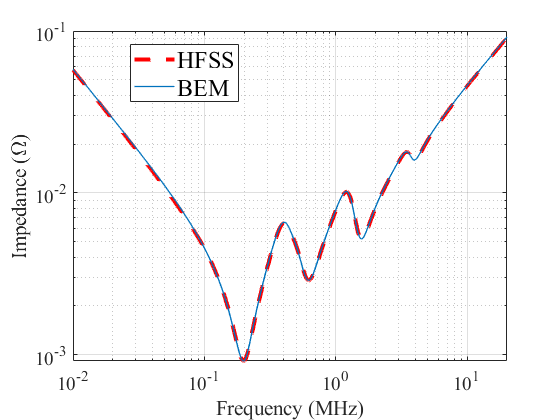}}
  \caption{}
  \label{fig2c}
\end{subfigure}
\caption{An irregular-shape power plane is used to verify the BEM algorithm \cite{[13]}. (a) Board shape and port distribution. The separation distance between each pair of power and ground vias is 2 mm. (b) Stackup. (c) Impedance comparison between BEM and HFSS simulation. \label{fig2}}
\end{figure}

Figure~\ref{fig2} demonstrates a test example \cite{[13]} to verify the BEM method by comparing it with an HFSS full-wave simulation \cite{[14]}. Figure~\ref{fig2a} describes the PCB shape. There are 6 ports formed by 6 pairs of power and ground vias. Port 6 is the observation port, and ports 1-5 are connected to decaps of 330 uF, 47uF, 10uF, 10uF, and 2.2uF respectively in Table~\ref{tab1}. Table~\ref{tab1} lists 10 different decap types represented by number 1-10 that will be used throughout this paper. Figure~\ref{fig2b} shows the stackup of this test board. Figure~\ref{fig2c} plots the results of the BEM method and the HFSS simulation. The observation frequency is from 0.01 MHz to 20 MHz. The perfect agreement in Figure~\ref{fig2c} strongly corroborates the accuracy and reliability of the BEM method. The BEM method, however, only consumes about 5 seconds, while the HFSS simulation spends over 5 minutes.

\begin{center}
\begin{table*}[htb]%
\caption{Decap Parameters \label{tab1}}
\centering
\begin{tabularx}{0.6\textwidth} { 
  | >{\centering\arraybackslash}X
  | >{\centering\arraybackslash}X 
  | >{\centering\arraybackslash}X
  | >{\centering\arraybackslash}X | }
 \hline
\textbf{Decap number} & \textbf{Capacitance ($\mu$F)}  & \textbf{ESL (nH)} & \textbf{ESR (m$\Omega$)} \\
 \hline
 1 & 0.1 & 0.19 & 34.7 \\ 
 2 & 0.47 & 0.18 & 18.3 \\ 
 3 & 1 & 0.22 & 15.2 \\ 
 4 & 2.2 & 0.20 & 7.2 \\ 
 5 & 4.7 & 0.28 & 7.1 \\ 
 6 & 10 & 0.26 & 5.2 \\ 
 7 & 22 & 0.27 & 4.0 \\ 
 8 & 47 & 0.15 & 2.9 \\ 
 9 & 220 & 0.41 & 1.9 \\ 
 10 & 330 & 0.46 & 1.2 \\ 
 \hline
\end{tabularx}
\end{table*}
\end{center}

\subsection{Data Generation}

\begin{figure}[htb]
\begin{subfigure}{.5\textwidth}
  \centerline{\includegraphics[height=160pt]{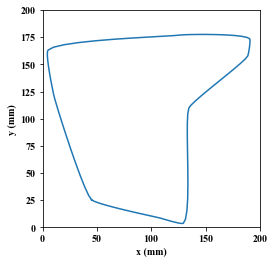}}
  \caption{}
  \label{fig3a}
\end{subfigure}
\begin{subfigure}{.5\textwidth}
  \centerline{\includegraphics[height=160pt]{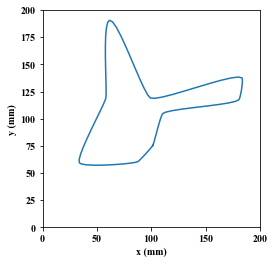}}
  \caption{}
  \label{fig3b}
\end{subfigure}
\caption{Two examples of randomly generated shapes. The maximum board size is 200mm × 200mm. \label{fig3}}
\end{figure}

To mimic different possible board shapes in real PCB designs, an algorithm \cite{[15]} was adopted to generate random 2D shapes. First, the maximum board size is specified as 200mm × 200mm. Then, the algorithm generates several random points (8 points are used in this paper) within the constrained board area. The generated points are sorted along one rotational direction and connected smoothly to form a closed contour. Figure~\ref{fig3} shows 2 randomly generated 2D shapes using the method.

\begin{figure}[htb]
\begin{subfigure}{.5\textwidth}
  \centerline{\includegraphics[height=160pt]{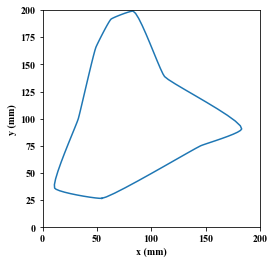}}
  \caption{}
  \label{fig4a}
\end{subfigure}
\begin{subfigure}{.5\textwidth}
  \centerline{\includegraphics[height=133pt]{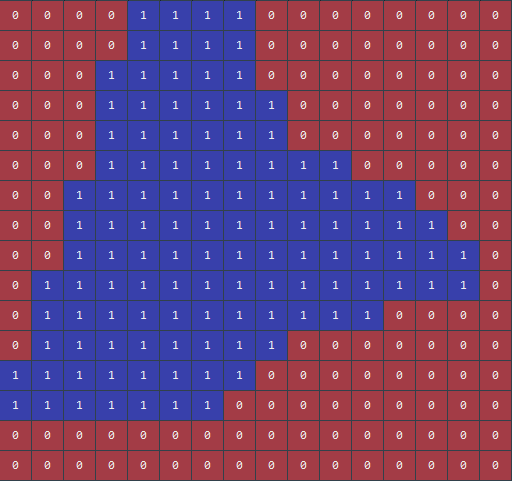}}
  \caption{}
  \label{fig4b}
\end{subfigure}
\caption{(a) An example of a randomly generated board shape. (b) The matrix representation of the board shape using a matrix of 16 × 16. \label{fig4}}
\end{figure}

For machine learning applications, input parameters need to be encoded into matrices. In this paper, a 2D matrix is used to represent the board shape.  Figure~\ref{fig4} illustrates an example of encoding and approximating a randomly generated board shape into a matrix of 16 × 16. The same dimension will be used for the remainder of this paper. We assume that each unit cell in the 16 × 16 board matrix can only contain either one IC port or one decap port. Moreover, each decap port is assumed to be horizontally oriented (along x-direction), and the distance between the power and ground vias is 2 mm. For simplification purposes, the IC port is also represented by a pair of power and ground vias that are 2 mm apart and horizontally oriented (along x-direction).

\begin{figure}[htb]
\begin{subfigure}{.5\textwidth}
  \centerline{\includegraphics[height=160pt]{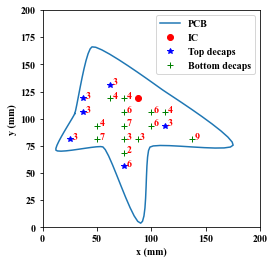}}
  \caption{}
  \label{fig5a}
\end{subfigure}
\begin{subfigure}{.5\textwidth}
  \centerline{\includegraphics[height=133pt]{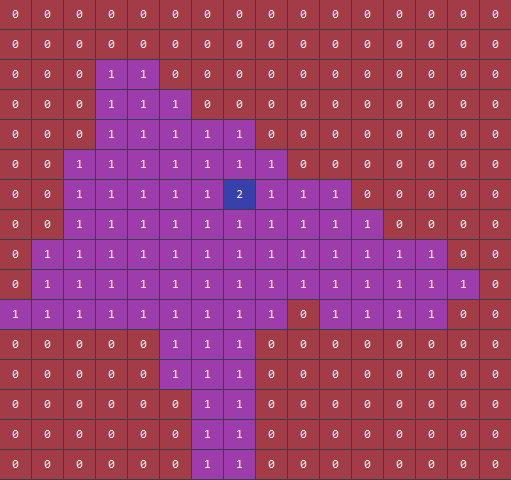}}
  \caption{}
  \label{fig5b}
\end{subfigure}
\begin{subfigure}{.5\textwidth}
  \centerline{\includegraphics[height=133pt]{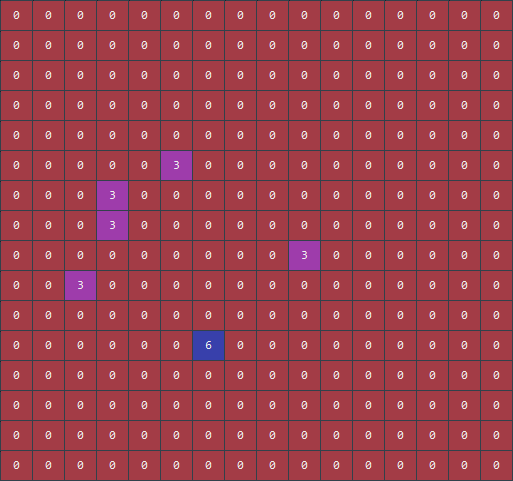}}
  \caption{}
  \label{fig5c}
\end{subfigure}
\begin{subfigure}{.5\textwidth}
  \centerline{\includegraphics[height=133pt]{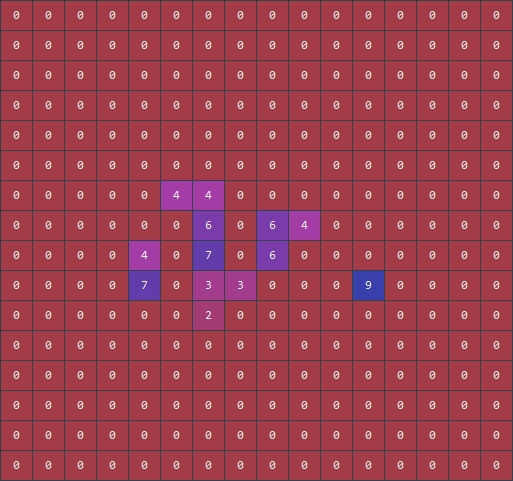}}
  \caption{}
  \label{fig5d}
\end{subfigure}
\caption{(a) An example of randomly generated board shape, IC location, and decap locations on the top and bottom side. The numbers represent the placed decaps corresponding to TABLE I. (b) The matrix representation of the board shape and the IC location using a matrix of 16×16; number 1 represents the board shape, and number 2 represents the IC location. (c) The matrix representation of the top decaps using a matrix of 16×16. (d) The matrix representation of the bottom decaps using a matrix of 16×16. \label{fig5}}
\end{figure}

To consider the variations of IC location and decap placement, different possible combinations are generated randomly inside the PCB area. The number of decaps is a random value from 1 to 19, and they are randomly distributed on the top and bottom layers. The IC port is also randomly located on the top layer. Each decap port is connected to a decap randomly chosen from Table~\ref{tab1} and denoted by a number from 1 to 10.  Figure~\ref{fig5a} shows an example with random IC and decap distributions. Three 16×16 matrices are used to describe the board shape, IC location, top decap placement, and bottom decap placement. The first matrix, as shown in Figure~\ref{fig5b}, defines the board shape and IC location using 1 and 2 respectively. The second and the third matrix represent the top and the bottom decap placement respectively, as shown in Figure~\ref{fig5b} and Figure~\ref{fig5c}. These three matrices are cascaded into a 3×16×16 matrix that will be used as the first input matrix of the DNN.

\begin{figure}[htb]
\begin{subfigure}{.5\textwidth}
  \centerline{\includegraphics[height=120pt]{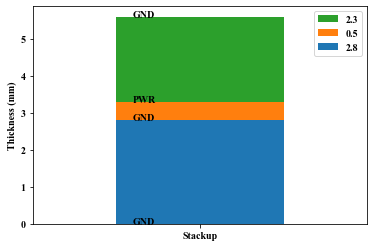}}
  \caption{}
  \label{fig6a}
\end{subfigure}
\begin{subfigure}{.5\textwidth}
  \centerline{\includegraphics[height=120pt]{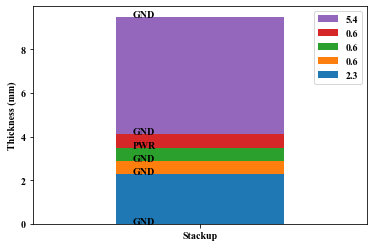}}
  \caption{}
  \label{fig6b}
\end{subfigure}
\caption{Two examples of randomly generated stackup. (a) 4 layers. (b) 6 layers. \label{fig6}}
\end{figure}

Another parameter to be included is the PCB stackup. A random stackup can be simply generated with a random thickness from 1 to 10 mm and a random number of layers from 4 to 9. The power layer is randomly located among the generated layers but cannot be located on the top layer or the bottom layer. The minimum distance between two adjacent layers is specified as 0.1 mm. Figure~\ref{fig6} shows two examples of randomly generated stackup, including the layer type and the dielectric thickness. In this paper, the relative permittivity of the PCB dielectric material is defined as 4.4.

\begin{figure}[htb]
\begin{subfigure}{1\textwidth}
  \centerline{\includegraphics[width=300pt,height=37.5pt]{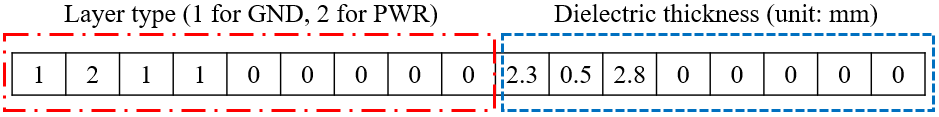}}
  \caption{}
  \label{fig7a}
\end{subfigure}

\begin{subfigure}{1\textwidth}
  \centerline{\includegraphics[width=300pt,height=37.5pt]{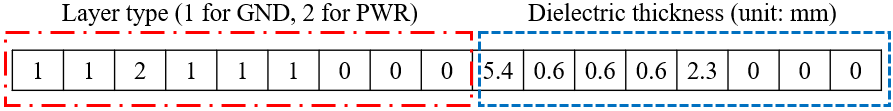}}
  \caption{}
  \label{fig7b}
\end{subfigure}
\caption{(a) The matrix representation of the stackup as shown in Figure~\ref{fig6a} using a 1D matrix of 1×17. (b) The matrix representation of the stackup as shown inFigure~\ref{fig6b} using a 1D matrix of 1×17. \label{fig7}}
\end{figure}

Similarly, the stackup information needs to be encoded into a matrix. Since using a 2D matrix is unnecessary, a 1D matrix of 1×17 is used instead. Since the maximum number of layers is 9, the first 9 elements of the 1×17 matrix define the layer type, in which 1 means ground layer, 2 means power layer, and 0 means empty (number 0 only appears when the number of layers is less than 9). The last 8 elements of the 1×17 matrix represent the dielectric thickness in millimeters, in which 0 also means empty. This stackup matrix will be the second input matrix of the DNN. Figure~\ref{fig7} shows the matrix form for the two stackup examples in Figure~\ref{fig6}.

\section{DNN Training}\label{sec3}

\begin{figure}[htb]
\centerline{\includegraphics[width=500pt]{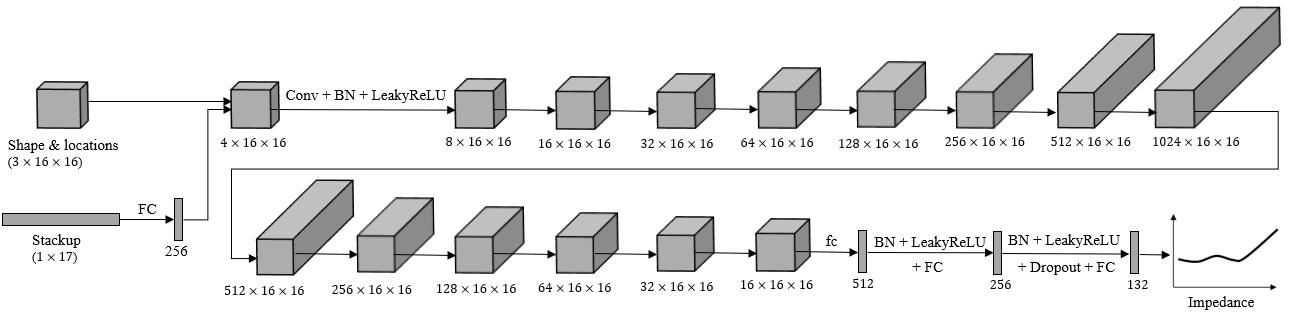}}
\caption{The detailed structure of the convolutional neural network (CNN). \label{fig8}}
\end{figure}

As introduced earlier, there are two input matrices for the DNN – the first 3×16×16 matrix defines the board shape, IC location, and decap placement, and the second 1×17 matrix defines the stackup information. These two matrices have different dimensions and should be combined into one image-like matrix so that the convolutional neural network (CNN) \cite{[6]} can be applied. To achieve this purpose, a fully connected (FC) layer is used to convert the 1×17 matrix to a 1×256 matrix, which is further reshaped to a 16×16 matrix and cascaded with the 3×16×16 matrix. Thus a 4×16×16 matrix is formed that can be followed by a series of convolutional layers. The detailed structure of the CNN is depicted in Figure~\ref{fig8}.

Starting from the 4×16×16 matrix, 14 convolutional layers are connected in series. In each convolutional layer, the kernel size is 3, the padding size is 1, and the stride is 1. Also, each convolutional layer is followed by a batch normalization (BN) layer \cite{[16]} and a Leaky ReLU activation layer \cite{[17]}. After the convolutional layers, several FC layers are utilized to reduce the matrix size to 132, which is the size of the output impedance matrix. A dropout layer \cite{[18]} is applied between the last two FC layers to prevent overfitting. 

\begin{figure}[htb]
\centerline{\includegraphics[width=210pt]{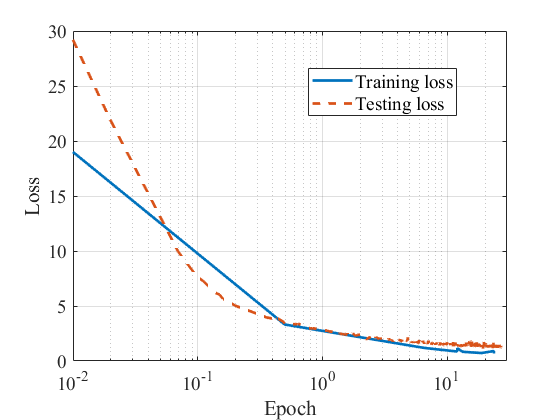}}
\caption{The convergence of the training loss and testing loss during the training process. \label{fig9}}
\end{figure}

By adopting the method of generating random board configurations, 13,000 PCBs with different IC locations, decap locations, and stackups were randomly generated. For each of these PCBs, the maximum number of decap locations was 19, and the BEM and the node voltage method were applied to calculate the Z-parameters. These Z-parameters were used repeatedly to connect with different decap combinations. For each PCB, 100 different decap placement scenarios, with different decap number (0 to 19) and different decap values (1 to 10), were randomly created, for a total number of 1.3 million groups of data. The entire data generation process took about one week. For each case, the decibel (dB) values of the impedance were used as the DNN output. The frequency range is from 10 kHz to 20 MHz.

Among all the generated data, 10,000 groups of data were used as a testing set, with the remaining used for training. The batch size was 128. The learning rate was 0.0001, and the Adam optimizer was utilized. The loss function was defined as the root mean square error (RMSE). One NVIDIA Tesla K80 GPU was used to accelerate the training. The training and the testing loss are plotted in Figure~\ref{fig9}. After 20 epochs, which took about 80 hours, both the training and testing loss converged stably to a low value close to 1, which indicates that the RMSE for the testing cases is only approximately 1 dB. 

\section{DNN Testing}\label{sec4}

\begin{figure}[htb]
\begin{subfigure}{.25\textwidth}
  \centerline{\includegraphics[width=125pt]{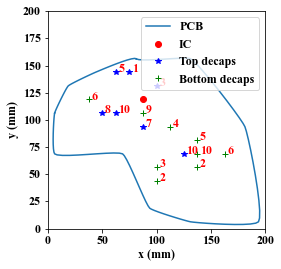}}
  \caption{}
  \label{fig10a}
\end{subfigure}
\begin{subfigure}{.35\textwidth}
  \centerline{\includegraphics[width=182pt]{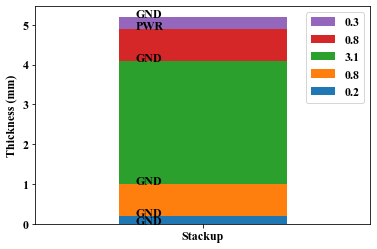}}
  \caption{}
  \label{fig10b}
\end{subfigure}
\begin{subfigure}{.4\textwidth}
  \centerline{\includegraphics[width=180pt]{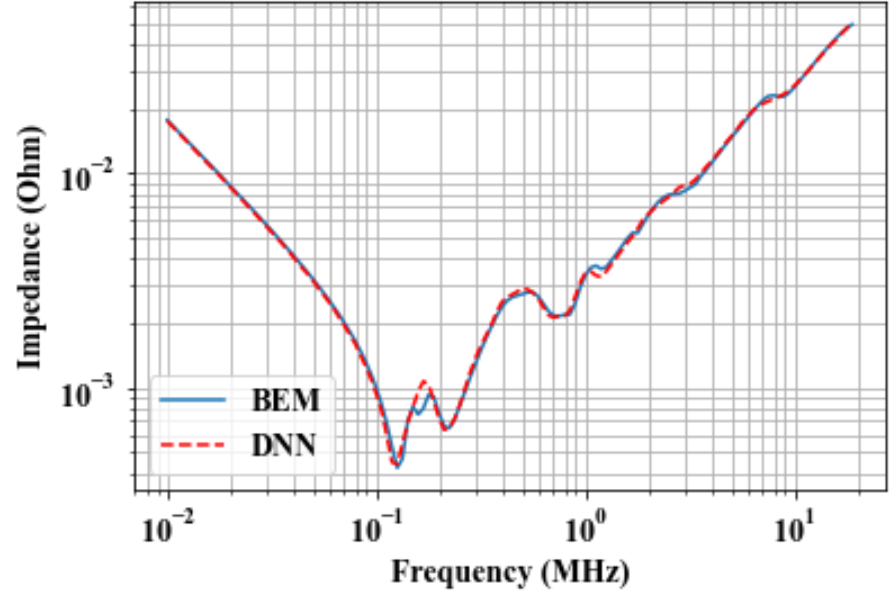}}
  \caption{}
  \label{fig10c}
\end{subfigure}
\caption{The first randomly selected test case. (a) PCB shape, IC location, and decap placement. (b) Stackup. (c) Comparison between the predicted impedance by the trained DNN and the calculated impedance by the BEM method. \label{fig10}}
\end{figure}

\begin{figure}[htb]
\begin{subfigure}{.25\textwidth}
  \centerline{\includegraphics[width=125pt]{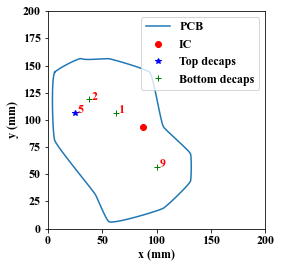}}
  \caption{}
  \label{fig11a}
\end{subfigure}
\begin{subfigure}{.35\textwidth}
  \centerline{\includegraphics[width=182pt]{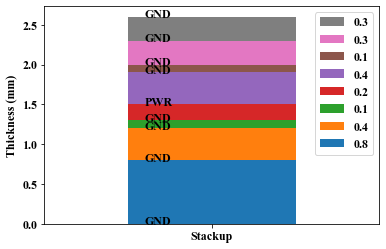}}
  \caption{}
  \label{fig11b}
\end{subfigure}
\begin{subfigure}{.4\textwidth}
  \centerline{\includegraphics[width=180pt]{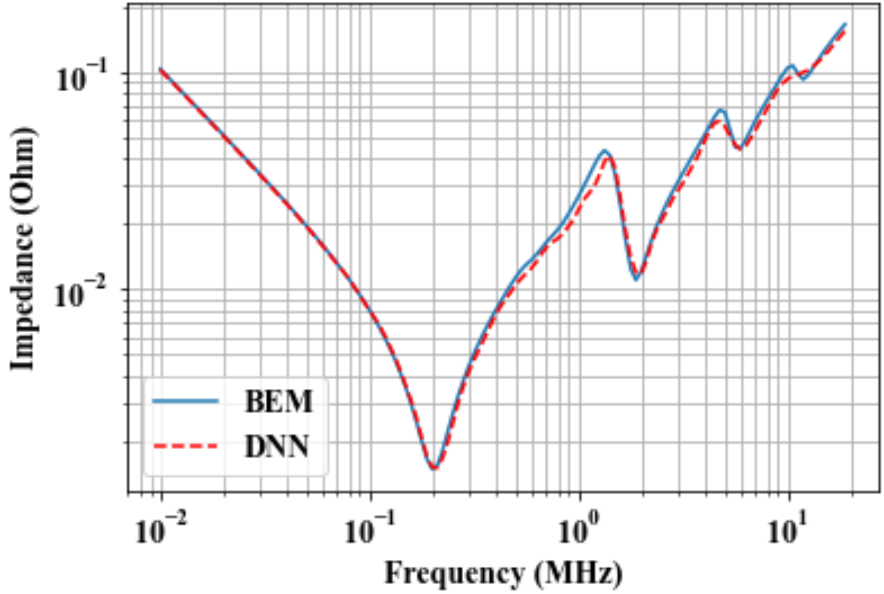}}
  \caption{}
  \label{fig11c}
\end{subfigure}
\caption{The second randomly selected test case. (a) PCB shape, IC location, and decap placement. (b) Stackup. (c) Comparison between the predicted impedance by the trained DNN and the calculated impedance by the BEM method. \label{fig11}}
\end{figure}

The trained DNN has a low testing loss as seen from Figure~\ref{fig9}. To further validate how the trained DNN behaves in predicting the impedance curve, two test cases are randomly selected from the testing dataset. The validation results of these two cases are shown in Figure~\ref{fig10} and Figure~\ref{fig11}. The impedance curves predicted by the trained DNN have a good agreement with the calculated curves by the BEM method. Using full-wave commercial products to simulate the impedance for similar structures requires more than 10 minutes. The BEM method reduces the computation time for these two cases to 10 seconds and 30 seconds respectively. The trained DNN, however, only needs 0.1 seconds for both cases on a normal CPU, which is hundreds of times faster than the BEM method and thousands of times faster than full-wave simulations. The detailed time comparison is listed in Table~\ref{tab2}. 

\begin{center}
\begin{table}[htb]%
\caption{Time Comparison \label{tab2}}
\centering
\begin{tabularx}{0.6\textwidth} { 
  | >{\centering\arraybackslash}X
  | >{\centering\arraybackslash}X 
  | >{\centering\arraybackslash}X | }
 \hline
\textbf{Methods} & \textbf{Case 1}  & \textbf{Case 2}  \\
 \hline
 Full-wave simulation & > 10 min & > 10 min \\ 
BEM & 10 s & 30 s \\ 
DNN & 0.1 s & 0.1 s \\ 
 \hline
\end{tabularx}
\end{table}
\end{center}

\section{Conclusions}\label{sec5}

In this paper, a novel concept of using deep learning to predict PDN impedance while considering the variations of board shape, stackup, IC location, and decap placement is proposed. A boundary element method (BEM) and the well-known node voltage method are adopted to quickly calculate the PDN impedance for arbitrary board shapes and stackup, which allows the algorithm to generate 1.3 million groups of training data with different board shapes, stackup, IC location, and decap placement. A convolution neural network (CNN) is constructed and trained with the produced data. The trained CNN can predict the impedance accurately for the testing cases, with a root mean square error (RMSE) of around 1 dB only. But the trained CNN has a much faster prediction speed than both full-wave simulations and the BEM method, using only 0.1 seconds. This deep learning algorithm can be a powerful tool for the application scenarios where a super-fast PDN impedance estimation is demanded. 


\section*{Acknowledgments}
This paper is based upon work supported by the National Science Foundation under Grant No. IIP-1916535 and a Google Faculty Research Award. 

\nocite{*}
\bibliography{wileyNJD-AMA}%

\vbox{}
\vbox{}
\vbox{}
\vbox{}
\vbox{}
\vbox{}
\vbox{}

\section*{Author Biography}

\begin{biography}{\includegraphics[height=70pt]{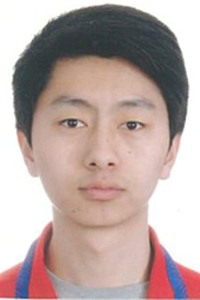}}{\textbf{Ling Zhang} received the B.S. degree in electrical engineering from Huazhong University of Science and Technology, Wuhan, China in June 2015, and MS degree from Missouri University of Science and Technology in Dec 2017. He worked in Cisco as a student intern from Aug 2016 to Aug 2017. He is now pursuing his PhD degree in EMC lab, Missouri University of Science and Technology. His research interests include machine learning applications in PI/SI/EMC, RFI source reconstruction and coupling paths analysis, and Emission Source Microscopy (ESM) technique for EMI source localization.}
\end{biography}

\vbox{}

\begin{biography}{\includegraphics[height=70pt]{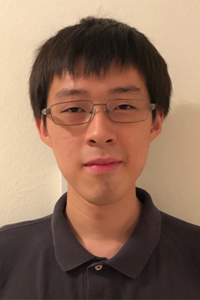}}{\textbf{Jack Juang} Jack Juang is a MS Degree student with the EMC Laboratory at Missouri University of Science and Technology (S\&T). He received his BS degree in Electrical Engineering from S\&T in 2020. His research interests include power distribution network modelling and optimization. He has been involved in projects relating to the optimization of decoupling capacitor placement and RF susceptibility of smart devices.}
\end{biography}

\vbox{}

\begin{biography}{\includegraphics[height=70pt]{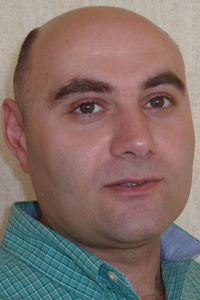}}{\textbf{Zurab Kiguradze} received the B.S., M.S., and Ph.D. degrees in mathematics from Tbilisi State University, Tbilisi, Georgia, in 1997, 1999, and 2003, respectively.
From 1999 to 2018, he was with Tbilisi State University. From 2015-2018 he was with Georgian Technical University. He is currently a Visiting Research Associate Professor at the Missouri University of Science and Technology, Rolla, MO, USA. His research interests include applied mathematics, algorithms, machine learning, mathematical modeling, optimization methods, development of mathematical algorithms for signal and power integrity.}
\end{biography}

\vbox{}

\begin{biography}{\includegraphics[height=70pt]{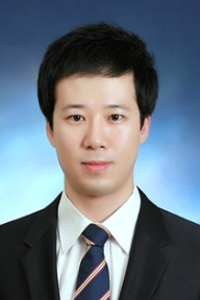}}{\textbf{Bo Pu} received the B.S. degree in electrical engineering from the Harbin Institute of Technology, China, in 2009, and the Ph.D. degree in electronic and electrical engineering from Sungkyunkwan University, South Korea, in 2015. From 2015 to 2020, he was a Staff Engineer in Foundry Business, Semiconductor R\&D Division of Samsung Electronics, Hwaseong, Korea. In July 2020, He joined the Missouri University of Science and Technology (formerly University of Missouri-Rolla), where he is currently a visiting assistant research professor of National Science Foundation (NSF) Industry/University Cooperative Research Center (I/UCRC) for Electromagnetic Compatibility.
His research interests include the design methodology for chip-package-PCB systems in areas of signal/power integrity and EMC. He recently focuses on the researches of high-speed integrated circuits system up to 224 Gbps, 2.5D Si-interposer for high bandwidth memory (HBM), and through silicon via (TSV) for 3-D ICs. He holds 10 patents about high speed links and 2.5D/3D ICs. Dr. Pu was the recipient of Best Student Paper Award as first author at the IEEE APEMC 2011, a Young Scientists Award from the International Union of Radio Science (URSI) in 2014, and the 2019 Distinguish reviewer for IEEE Transactions on Electromagnetic Compatibility. He also obtained Ph.D. Fellowship Award in 2013, Best Innovation Award, Excellent Performance Award, and Excellent Project Award as the first awardee in 2015-2019 from Samsung Electronics. He was as a Session Chair in IEEE APEMC 2017, IEEE EMC+SIPI 2020, and a TPC Member of the Joint IEEE EMCS \& APEMC 2018. He is currently an Associate Editor for IEEE Access and moderator of IEEE TechRxiv as well as a senior member of IEEE.}
\end{biography}

\vbox{}

\begin{biography}{\includegraphics[height=70pt]{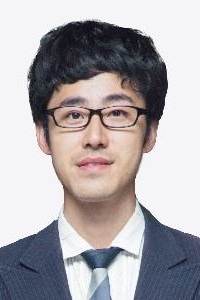}}{\textbf{Shuai Jin} received the B.S. degree in biomedical engineering from Huazhong University of Science and Technology, Wuhan, Hubei, China in 2011 and received the M.S. and Ph.D. degree in electrical engineering from the Missouri University of Science and Technology (University of Missouri–Rolla), Rolla, MO, in 2013 and 2017 respectively. He is currently working as signal integrity engineer in Google Platform. His research interests include signal integrity in high speed digital systems, power distributed network modeling, RF interference and high-speed package modeling.}
\end{biography}

\vbox{}

\begin{biography}{\includegraphics[height=70pt]{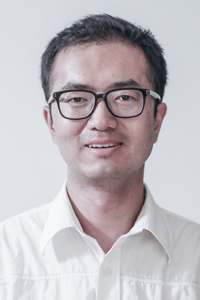}}{\textbf{Songping Wu} is a SI/PI/RF desense lead in Google ChromeOS team. Prior to joining Google, he was a senior SI engineer at Apple and a hardware engineer at Cisco. He has published more than 50 research papers and holds 6 patents. His research results and patents have been applied to Google ChromeBooks, Apple iPhones and Cisco UCS servers. He is an IEEE Senior Member and recipient of the 2011 IEEE EMC Society President's Memorial Award. He is the chair of the IEEE EMC Society TC-10 (Signal Integrity and Power Integrity). He obtained his Ph.D. degree from the Missouri University of Science and Technology and received the B.S. degree from Wuhan University, Wu-han, China, in 2003, the M.S. degree from the Huazhong University of Science and Technology, Wuhan, China, in 2006.}
\end{biography}

\vbox{}

\begin{biography}{\includegraphics[height=70pt]{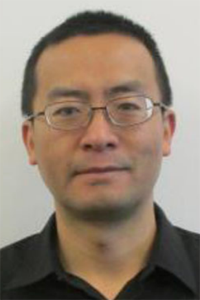}}{\textbf{Zhiping Yang} received the
B.S. and M.S. degrees from Tsinghua University, Beijing, China, in 1994 and 1997, respectively, and the Ph.D. degree from the University of MissouriRolla, Rolla, MO, USA, in 2000, all in electrical engineering. From 2000 to 2005, he was a Technical Leader with Cisco Systems, San Jose, CA, USA. From 2005 to 2006, he was a Principal Engineer with Apple
Computer, Cupertino, CA, USA. From 2006 to 2012, he worked in Nuova Systems (which was acquired by Cisco in 2008) and Cisco Systems, San Jose, CA, as a Principal Engineer. From
2012 to 2015, he was a Senior Manager with Apple, Cupertino, CA, USA. He is currently a Senior Hardware Manager with Google Consumer Hardware Group, Mountain View, CA, USA. He has authored or coauthored more than 40 research papers and 17 patents. His research and patents have been applied in Google Chromebook, Apple iPhone 5S/6/6S, Cisco UCS, Cisco Nexus 6K/4K/3K, and Cisco Cat6K products. His current research interests include signal integrity and power integrity methodology development for Die/Package/Board co-design,
high-speed optical module, various high-speed cabling solutions, high-speed DRAM/storage technology, high-speed serial signaling technology, and RF interference.
Dr. Yang was the recipient of the 2016 IEEE EMCS Technical Achievement Award.}
\end{biography}

\vbox{}

\begin{biography}{\includegraphics[height=70pt]{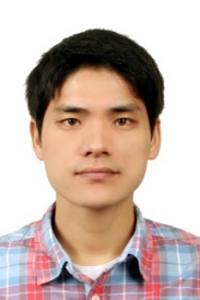}}{\textbf{Chulsoon Hwang} is with EMC laboratory at Missouri S\&T (formerly University of Missouri-Rolla) as an assistant professor. He received his PhD degree from KAIST, Daejeon, South Korea in 2012 and worked at Samsung Electronics as a senior engineer from 2012 to 2015. His research interests include signal/power integrity in high-speed digital systems, RF desensitization, EMI, hardware security and machine learning.
Dr. Hwang received the Young Science Award at AP-EMC 2018, the 2019 Google Faculty Research Award, and a best paper award at AP-EMC 2017, IEEE EMC+SIPI 2019, DesignCon 2018 and 2019. }
\end{biography}

\end{document}